\documentclass{article}
\usepackage{spconf,amsmath,graphicx}
\usepackage{times}
\usepackage{latexsym}
\usepackage{amsfonts}
\usepackage{multirow}
\usepackage{graphicx}
\usepackage{amssymb}
\usepackage{amsmath}
\usepackage{url}

\title{Dynamic Time-Aware Attention to Speaker Roles and Contexts \\for Spoken Language Understanding}
\name{Po-Chun Chen$^\star$\quad Ta-Chung Chi$^\star$\quad Shang-Yu Su$^\dagger$\quad Yun-Nung Chen$^\star$\thanks{The first three authors have equal contributions.}}
\address{$^\star$Department of Computer Science and Information Engineering\\$^\dagger$Graduate Institute of Electrical Engineering\\National Taiwan University, Taipei, Taiwan\\\texttt{ \small \{b02902019,r06922028,r05921117\}@ntu.edu.tw\quad y.v.chen@ieee.org}}
\begin{document}
%
\maketitle
\begin{abstract}
Spoken language understanding (SLU) is an essential component in conversational systems.
Most SLU components treat each utterance independently, and then the following components aggregate the multi-turn information in the separate phases.
In order to avoid error propagation and effectively utilize contexts, prior works leveraged history for contextual SLU.
However, the previous models only paid attention to the content in history utterances without considering their temporal information and speaker roles.
In dialogues, the most recent utterances should be more important than the least recent ones.
Furthermore, users usually pay attention to 1) self history for reasoning and 2) others’ utterances for listening, the speaker of the utterances may provides informative cues to help understanding.
Therefore, this paper proposes an attention-based network that additionally leverages temporal information and speaker role for better SLU, where the attention to contexts and speaker roles can be automatically learned in an end-to-end manner.
The experiments on the benchmark Dialogue State Tracking Challenge 4 (DSTC4) dataset show that the time-aware dynamic role attention networks significantly improve the understanding performance\footnote{The released code: \url{https://github.com/MiuLab/Time-SLU}}.
\end{abstract}
\begin{keywords}
dialogue, language understanding, SLU, temporal, role, attention, deep learning.
\end{keywords}

\section{Introduction}
\label{sec:intro}
Spoken dialogue systems that can help users to solve complex tasks such as booking a movie ticket have become an emerging research topic in artificial intelligence and natural language processing area. 
With a well-designed dialogue system as an intelligent personal assistant, people can accomplish certain tasks more easily via natural language interactions. 
Today, there are several virtual intelligent assistants, such as Apple's Siri, Google's Home, Microsoft's Cortana, and Amazon's Echo. Recent advance of deep learning has inspired many applications of neural models to dialogue systems. Prior work introduced network-based end-to-end trainable task-oriented dialogue systems~\cite{wen2017network,bordes2017learning,dhingra2017towards,li2017end}.

A key component of the understanding system is a spoken language understanding (SLU) module---it parses user utterances into semantic frames that capture the core meaning, where three main tasks of SLU are domain classification, intent determination, and slot filling~\cite{tur2011spoken}.
A typical pipeline of SLU is to first decide the domain given the input utterance, and based on the domain, to predict the intent and to fill associated slots corresponding to a domain-specific semantic template.
With the power of deep learning, there are emerging better approaches of SLU~\cite{hakkani2016multi,chen2016knowledge,chen2016syntax,wang2016learning}.
However, the above work focused on single-turn interactions, where each utterance is treated independently.

\begin{figure*}[t]
\centering
\includegraphics[width=\linewidth]{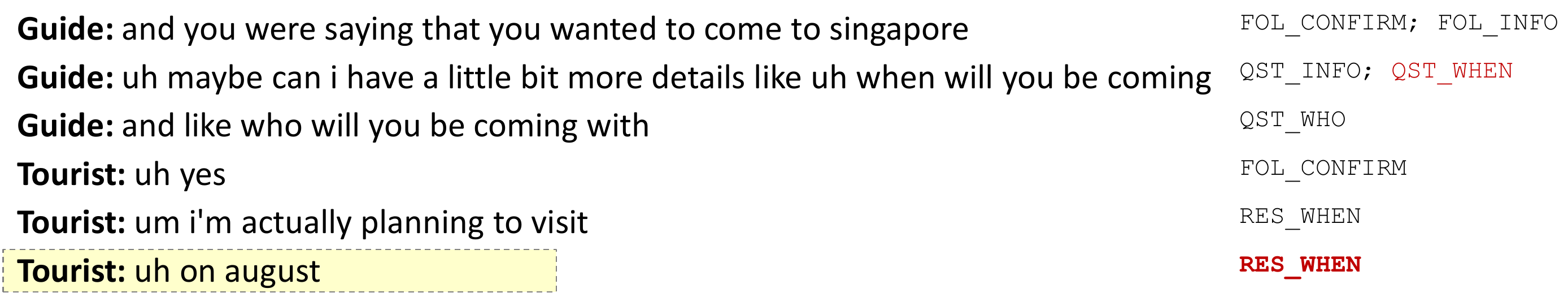}
\caption{The human-human conversational utterances and their associated semantic labels from DSTC4.}
\label{fig:example}
\end{figure*}

The contextual information has been shown useful for SLU~\cite{bhargava2013easy,xu2014contextual,chen2015leveraging,sun2016an}.
For example, Figure~\ref{fig:example} shows conversational utterances, where the intent of the highlighted tourist utterance is to ask about location information, but it is difficult to understand without contexts.
Hence, it is more likely to estimate the location-related intent given the contextual utterances about location recommendation.
Contextual information has been incorporated into the recurrent neural network (RNN) for improved domain classification, intent prediction, and slot filling~\cite{xu2014contextual,shi2015contextual,weston2015memory,chen2016end}.



Most of previous dialogue systems did not take speaker role into consideration.
However, different speaker roles can cause notable variance in speaking habits and later affect the system performance differently. 
From Figure~\ref{fig:example}, the benchmark dialogue dataset, Dialogue State Tracking Challenge 4 (DSTC4)~\cite{kim2016fourth}\footnote{\url{http://www.colips.org/workshop/dstc4/}}, contains two specific roles, a tourist and a guide.
Under the scenario of dialogue systems and the communication patterns, we take the tourist as a user and the guide as the dialogue agent (system).
During conversations, the user may focus on not only \emph{reasoning (user history)} but also \emph{listening (agent history)}, so different speaker roles could provide various cues for better understanding~\cite{chi2017speaker}.

In addition, neural models incorporating attention mechanisms have had great successes in machine translation~\cite{bahdanau2014neural}, image captioning~\cite{xu2015show}, and various tasks. 
Attentional models have been successful because they separate two different concerns: 1) deciding which input contexts are most relevant to the output and 2) actually predicting an output given the most relevant inputs. 
For example, the highlighted current utterance from the tourist, ``\textit{uh on august}'', in the conversation of Figure~\ref{fig:example} is to respond the question about \textit{when}, and the content-aware contexts that can help current understanding are the first two utterances from the guide ``\textit{and you were saying that you wanted to come to singapore}'' and ``\textit{un maybe can i have a little bit more details like uh when will you be coming}''.
Although content-aware contexts may help understanding, the most recent contexts may be more important than others.
In the same example, the second utterance is more related to the \textit{when} question, so the temporal information can provide additional cues for the attention design.

This paper focuses on investigating various attention mechanism in neural models with contextual information and speaker role modeling for language understanding. 
In order to comprehend what tourist is talking about and imitate how guide react to these meanings, this work proposes a role-based contextual model by modeling role-specific contexts differently for improving the system performance and further design associated time-aware and content-aware attention mechanisms.

\begin{figure*}[t]
\centering
\includegraphics[width=\linewidth]{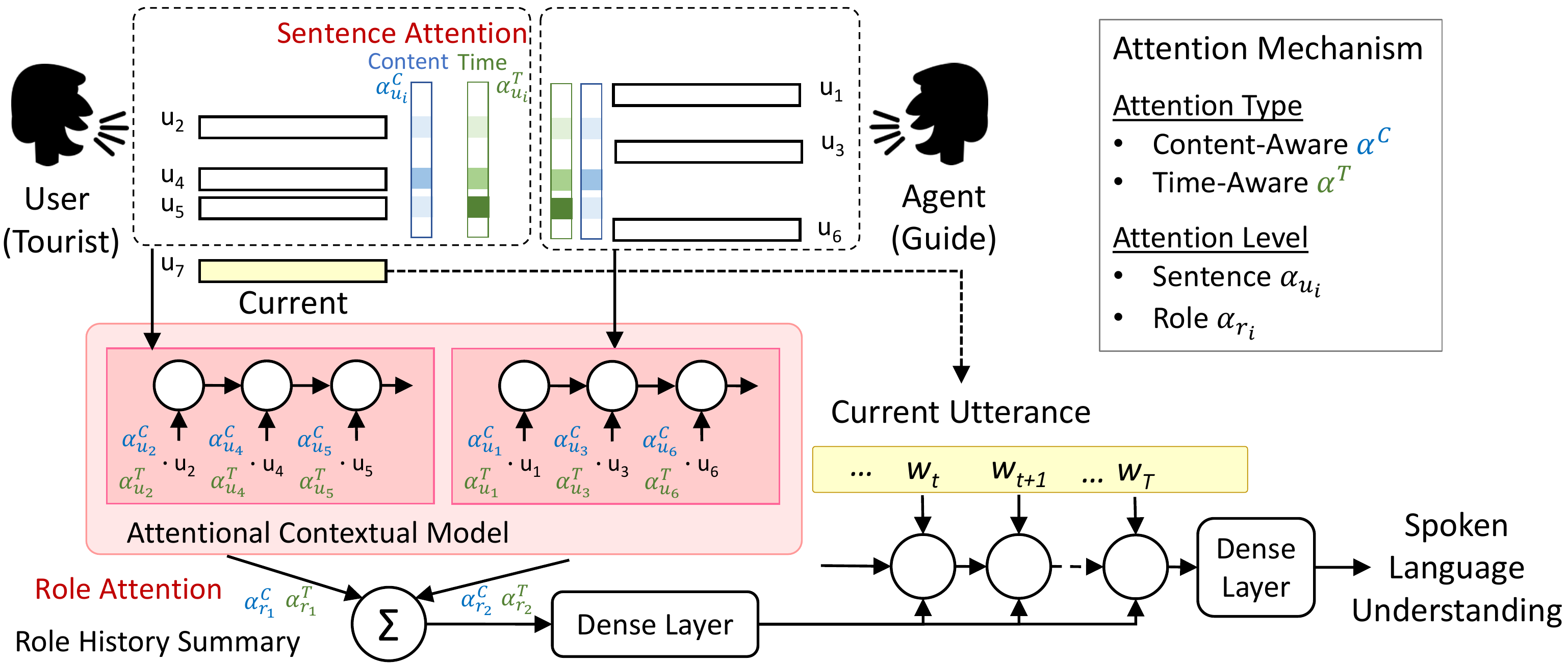}
\caption{Illustration of the proposed attentional contextual model.}
\label{fig:model}
\end{figure*}

\section{Proposed Approach}
\label{sec:model}

The model architecture is illustrated in Figure~\ref{fig:model}.
First, the previous utterances are fed into the contextual model to encode into the history summary, and then the summary vector and the current utterance are integrated for helping understanding.
The contextual model leverages the attention mechanisms illustrated in the red block, which implements different attention types and attention levels. 
The whole model is trained in an end-to-end fashion, where the history summary vector and the attention weights are automatically learned based on the downstream SLU task.
The objective of the proposed model is to optimize the conditional probability $p(\mathbf{\hat{y}}\mid \mathbf{x})$, so that the difference between the predicted distribution and the target distribution, $q(y_k=z\mid \mathbf{x})$, can be minimized:
\begin{equation}
\mathcal{L}=-\sum_{n=1}^{N}\sum_{k=1}^{K}q(y_k=z\mid \mathbf{x}) \log p(\hat{y_k}=z\mid \mathbf{x}),
\end{equation}
where $n$ is the number of samples and the labels $\textbf{y}$ are the labeled intent tags for understanding.

\subsection{Contextual Language Understanding }
\label{ssec:clu}
Given the current utterance $\textbf{x}=\{w_t\}^T_1$, the goal is to predict the user intents of $\textbf{x}$, which includes the speech acts and associated attributes shown in Figure~\ref{fig:example}.
We apply a bidirectional long short-term memory (BLSTM) model~\cite{schuster1997bidirectional} to integrate preceding and following words to learn the probability distribution of the user intents.
\begin{eqnarray}
\label{eq:basic}
\textbf{v}_\text{cur} &=& \text{BLSTM}(\textbf{x}, W_\text{his}\cdot \textbf{v}_\text{his}),\\
\textbf{o} &=& \mathtt{sigmoid}(W_\text{SLU}\cdot \textbf{v}_\text{cur}),
\end{eqnarray}
where $W_\text{his}$ is a dense matrix and $\textbf{v}_\text{his}$ is the history summary vector, $\textbf{v}_\text{cur}$ is the context-aware vector of the current utterance encoded by the BLSTM, and $\textbf{o}$ is the intent distribution.
Note that this is a multi-label and multi-class classification, so the $\mathtt{sigmoid}$ function is employed for modeling the distribution after a dense layer.
The user intent labels $\textbf{y}$ are decided based on whether the value is higher than a threshold $\theta$ tuned by the development set.
\subsection{Speaker Role Contextual Module}
\label{ssec:contexualmodel}
In order to leverage the contextual information, we utilize the prior contexts from two roles to learn history summary representations, $\textbf{v}_\text{his}$ in (\ref{eq:basic}).
The illustration is shown in the left-right part of Figure~\ref{fig:example}.



In a dialogue, there are at least two roles communicating with each other, each individual has his/her own goal and speaking habit.
For example, the tourists have their own desired touring goals and the guides' goal is try to provide the sufficient touring information for suggestions and assistance.
Prior work usually ignored the speaker role information or only modeled a single speaker's history for various tasks~\cite{chen2016end,yang2017end}.
The performance may be degraded due to the possibly unstable and noisy input feature space.

To address this issue, this work proposes the role-based contextual model: instead of using only a single BLSTM model for the history, we utilize one individual contextual module for each speaker role.
Each role-dependent recurrent unit $\text{BLSTM}_{\text{role}_i}$ receives corresponding inputs, $x_{t,\text{role}_i}$, which includes multiple utterances $u_i$ ($i=[1, ..., t-1]$) preceding to the current utterance $u_t$ from the specific role, $\text{role}_i$, and have been processed by an encoder model.
\begin{eqnarray}
\label{eq:tag2}
\textbf{v}_\text{his} &=& \textbf{v}_{\text{his}, \text{role}_a} + \textbf{v}_{\text{his}, \text{role}_b}\\
&=& \text{BLSTM}_{\text{role}_a}(x_{t,\text{role}_a}) + \text{BLSTM}_{\text{role}_b}(x_{t,\text{role}_b}),\nonumber
\end{eqnarray}
where $x_{t, \text{role}_i}$ contains vectors after one-hot encoding that represent the annotated intent and the attribute features.
Note that this model requires the ground truth annotations for history utterances for training and testing.
Therefore, each role-based contextual module focuses on modeling role-dependent goals and speaking style, and $\textbf{v}_\text{cur}$ from (\ref{eq:basic}) would contain role-based contextual information.

\subsection{Neural Attention Mechanism}
One of the earliest work with a memory component applied to language processing is memory networks~\cite{weston2015memory,sukhbaatar2015end}, which encode mentioned facts into vectors and store them in the memory for question answering (QA).
The idea is to encode important knowledge and store it into memory for future usage with attention mechanisms.
Attention mechanisms allow neural network models to selectively pay attention to specific parts.
There are also various tasks showing the effectiveness of attention mechanisms~\cite{xiong2016dynamic,chen2016end}. 
This paper focuses on two attention types: content-aware ($\alpha^C$) and time-aware ($\alpha^T$), and two attention levels: sentence ($\alpha_{u_i}$) and role ($\alpha_{r_i}$) illustrated in Figure~\ref{fig:model}.
The following sections first describe content-aware and time-aware attention mechanism using the sentence-level structure, and the role-level attention is explained afterwards.
\subsubsection{Content-Aware Attention}
Given the utterance contexts, the model can learn the attention to decide where to focus more for better understanding the current utterance~\cite{weston2015memory,chen2016end}.
In the content-aware attention, it is intuitive to use the semantic relation between the current utterance representations and the history context vector as the attention weight, which
is used as a measurement of how much the model should focus on a specific preceding utterance:
\begin{equation}
\label{eq:att_c}
\alpha^C_{u_i} = \mathtt{softmax}(M_{\text{S}}(\textbf{v}_\text{cur} + \textbf{x}_i)),
\end{equation}
where $\alpha^C_{u_i}$ is the content-aware attention vector highlighted as blue texts in Figure~\ref{fig:model}, the $\mathtt{softmax}$ function is used to normalize the attention values of the history utterances, and $M_\text{S}$ is an MLP for learning the attention weight given the current representation $\textbf{v}_\text{cur}$ and $\textbf{x}_i$ is the vector of $u_i$ in the history. 

Then when computing the history summary vector, the BLSTM encoder additionally considers the corresponding attention weight for each history utterance in order to emphasize the content-related contexts.
(\ref{eq:tag2}) can be rewritten into
\begin{eqnarray}
\label{eq:his_att_c}
\textbf{v}_\text{his} = \sum_{i\in \{a,b\}} \text{BLSTM}_{\text{role}_i}(x_{t,\text{role}_i}, \{ \alpha^C_{u_j} \mid u_j \in \text{role}_i\}).
\end{eqnarray}

\subsubsection{Time-Aware Attention}
Intuitively, most recent utterances contain more relevant information; therefore we introduce time-aware attention mechanism which computes attention weights by the time of utterance occurrence explicitly.
We first define the time difference between the current utterance and the preceding sentence as $d(u_i)$ for each preceding utterance $u_i$, and then simply use its reciprocal as the attention value, $\alpha^T_{u_i}$:
\begin{equation}
\alpha^T_{u_i} = \frac{1}{d(u_i)}.
\end{equation}

Here the importance of a earlier history sentence would be considerably compressed.
For the sentence-level attention, before feeding into the contextual module, each history vector is weighted by its reciprocal of distance.


\begin{eqnarray}
\textbf{v}_\text{his} = \sum_{i\in \{a,b\}} \text{BLSTM}_{\text{role}_i}(x_{t,\text{role}_i}, \{ \alpha^C_{u_j}\cdot \alpha^T_{u_j} \mid u_j \in \text{role}_i\}).
\end{eqnarray}

\subsubsection{Dynamic Speaker Role Attention}

Switching to role-level attention, a dialogue is disassembled from a different perspective about which speaker's information is more important.
From (\ref{eq:tag2}), we have the role-dependent history summary representations, $\textbf{v}_{\text{his}, \text{role}_i}$. 
The role-level attention is to decide how much to address on different speaker roles' contexts in order to better understand the current utterance.
\begin{equation}
\alpha^C_{r_i} = \mathtt{softmax}(M_R(\textbf{v}_\text{cur} + \textbf{v}_{\text{his}, \text{role}_i})), 
\end{equation}
where $\alpha_{r_i}$ is the attention weight of the role $i$, and $M_R$ is the MLP for learning the role-level content-aware attention weights, while $M_S$ in (\ref{eq:att_c}) is at the sentence level.

In terms of time-aware attention, experimental results show that the importance of a speaker given the contexts can be approximated to the minimum distance among the speaker's utterances\footnote{Different settings (min, avg, ) were attempted.}.
\begin{equation}
\alpha^T_{r_i} = \min \frac{1}{d(u_j)},
\end{equation}
where $u_j$ includes all contextual utterances from the speaker $r_i$.

With content-aware or time-aware attention at the role level,
the sentence-level history result from (\ref{eq:his_att_c}) can be rewritten into
\begin{equation}
\textbf{v}_\text{his} = \alpha^{C/T}_{\text{role}_a} \cdot \textbf{v}_{\text{his}, \text{role}_a}
+ \alpha^{C/T}_{\text{role}_b}\cdot \textbf{v}_{\text{his}, \text{role}_b},
\end{equation}
for combining role-dependent history vectors with their attention weights.

\subsection{End-to-End Training}
The objective is to optimize SLU performance, predicting multiple speech acts and attributes described in \ref{ssec:clu}.
In the proposed model, all encoders, prediction models, and attention weights (except time-aware attention) can be automatically learned in an end-to-end manner.


\begin{table*}
\centering
  \begin{tabular}{ | l c l | c | c | c | }
    \hline
     \bf Model & & \bf Attention Level & \bf Tourist & \bf Guide & \bf ~~All~~ \\ \hline\hline
    Baseline & (a) & \emph{DSTC4-Best 1} & 52.1 & 61.2 & 57.8 \\
    & (b) & \emph{DSTC4-Best 2} & 51.1 & 67.4 & 61.4 \\
    & (c) & w/o context & 63.2 & 69.2 & 66.6 \\
    & (d) & w/ context w/o speaker role & 68.3 & 74.4 & 71.6 \\
    & (e) & w/ context w/ speaker role & 69.1 & 74.4 & 72.1\\\hline
    Content-Aware Attention & (f) & Sentence & 68.5 & 73.6 & 71.6\\
    & (g) & Role & 68.6 & 74.0 & 71.8\\
    & (h) & Both & 67.9 & 73.5 & 71.5\\\hline
    Time-Aware Attention & (i) & Sentence & \bf 70.5$^\dag$ & \bf 77.7$^\dag$  & \bf 74.6$^\dag$ \\
    & (j) & Role & 70.0$^\dag$  & 76.8$^\dag$  & 74.2$^\dag$ \\
    & (k) & Both & 70.4$^\dag$  & 77.3$^\dag$  & 74.5$^\dag$ \\\hline
    Content-Aware + Time-Aware Attention & (l) & Sentence & 69.7$^\dag$  & 76.7$^\dag$  & 74.0$^\dag$ \\
    & (m) & Role & \bf 70.1$^\dag$  & \bf 77.1$^\dag$  & \bf 74.1$^\dag$ \\
    & (n) & Both & 69.5$^\dag$  & 76.2$^\dag$  & 73.5$^\dag$ \\
    \hline
  \end{tabular}
  
\caption{Spoken language understanding performance reported on F-measure in DSTC4 (\%). $^\dag$ indicates the significant improvement compared to all baseline methods.}
\label{tab:res}
\end{table*}

\section{Experiments}
\label{sec:experiments}
To evaluate the effectiveness of the proposed model, we conduct the language understanding experiments on human-human conversational data.

\subsection{Setup}
\label{ssec:settings}
The experiments are conducted on DSTC4, which consists of 35 dialogue sessions on touristic information for Singapore collected from Skype calls between 3 tour guides and 35 tourists~\cite{kim2016fourth}. 
All recorded dialogues with the total length of 21 hours have been manually transcribed and annotated with speech acts and semantic labels at each turn level.
The speaker information (guide and tourist) is also provided.
Unlike previous DSTC series collected human-computer dialogues, 
human-human dialogues contain rich and complex human behaviors and bring much difficulty to all the tasks.
Given the fact that different speaker roles behave differently and longer contexts, DSTC4 is a suitable benchmark dataset for evaluation.

We choose the mini-batch \textit{adam} as the optimizer with the batch size of 256 examples.
The size of each hidden recurrent layer is 128.
We use pre-trained 200-dimensional word embeddings $GloVe$~\cite{pennington2014glove}.
We only apply 30 training epochs without any early stop approach. 
We focus on predicting multiple labels including intents and attributes, so the evaluation metric is average F1 score for balancing recall and precision in each utterance.
The experiments are shown in Table~\ref{tab:res}, where we report the average results over five runs. 

\subsection{Baseline Results}

The baselines (rows (a) and (b)) are two of the best participants of DSTC4 in IWSDS 2016~\cite{kim2016fourth} (best results for tourist and guide understanding respectively).
It is obvious that tourist intents are much more difficult than guide intents (most systems achieved higher than 60\% of F1 for guide intents but lower than 50\% for tourist intents), because the guides usually follow similar interaction patterns. 
In our experiments, we focus more on the tourist part, because SLU in a dialogue system is to understand the users, who ask for assistance.

The baseline (c) only takes the current utterance into account without any history information, and then applies a simple BLSTM for SLU. 
The baselines (d) and (e) leverage the semantic labels of contexts for learning history summary vectors, achieving the improvement compared to the results without contexts.
The baseline (e) that uses history information and conducts our role-based contextual model to capture role-specific information separately slightly improves the SLU performance for both tourist and guide, achieving 69.1\% and 74.4\% for tourist and guide understanding respectively.

\subsection{Content-Aware Attention}
For the content-aware attention, our model learns the importance of each utterance on both sentence-level and role-level in an end-to-end fashion.
The table shows that the learned attention for both sentence-level and role-level does not yield improvement compared to the baseline (e). 
The probable reason is that the content-aware importance can be handled by the BLSTM when producing the history summary, and the learned attention does not provide additional cues for improvement.

\subsection{Time-Aware Attention}
With the time-aware attention, using the reciprocal of distance as attention value significantly improves the performance to about 70\% (tourist) and about 77\% (guide) using either sentence-level or role-level attention. 
The reason that the performance among different levels of time-aware mechanisms are close may be that the closest utterance is capable of capturing the importance of the history.
Using the reciprocal of minimum distance among tourist/guide history as their attention weights in the role-level attention can effectively pay the correct attention to the salient information, achieving promising performance.

However, the table shows that the sentence-level attention is slightly better than role-level attention for both tourist and guide parts when considering time-aware attention.
The reason may be that the intents are very diverse and hence require the more precise focus on the content-related history in order to achieve correct understanding results. 
Also, the results using both sentence-level and role-level attention do not differ a lot from the results using only sentence-level attention.

In addition to the current setting for time-aware attention, other trends of temporal decay can be further investigated.
We leave this part as the future work.

\subsection{Content-Aware and Time-Aware Attention}
The proposed model includes two attention types, content-aware and time-aware, and we integrate both types into a single model to analyze whether their advantages can be combined.
The rows (l)-(n) are the results using both content-aware and time-aware attention weights, but the performance is similar to the time-aware results.
For the results using only role-level attention, combining content-aware and time-aware attention weights obtains the improvement (row (j) v.s row (m)). 

\subsection{Comparison between Sentence-Level and Role-Level}
\label{ssec:sent_role}
Among the experiments using either content-aware attention or time-aware attention or both, we introduce both sentence-level and role-level mechanisms for analysis. 
From Table~\ref{tab:res}, rows (i) and (l) show that the sentence-level attention can benefit the results even though we do not leverage the content-aware attention.
On the other hand, rows (j) and (m) show that the role-level attention requires content-related information in order to achieve better performance.
The reason is probably that the role-level attention is not precise enough to capture the salience of utterances.
Therefore, combining with content-aware cues can effectively focus on the correct part.


\subsection{Dynamic Speaker Role Attention Analysis}
To further analyze the dynamic role-level attention, we compute the mean of the learned attention weights.
The results are shown in Table~\ref{tab:role_att}.

\begin{table}
\centering
\begin{tabular}{ | c | c | c | c | c | c | }
    \hline
    \multirow{2}{*}{Task} & \multicolumn{2}{|c|}{Role-Level Attention Weight}\\
    \cline{2-3}
 & Tourist Context & Guide Context\\
\hline \hline
Tourist Understanding & 0.48 & 0.52\\
Guide Understanding & 0.30 & 0.70\\
    \hline
  \end{tabular}
\caption{The average role-level attention weight learned from the proposed model using the role-level attention mechanism.}
\label{tab:role_att}
\end{table}

For tourist understanding, the tourist and the guide attention weights are 0.48 and 0.52 respectively. 
We can therefore conclude that both tourist and guide history are likewise important in order to understand the tourist utterances.
The observation matches our analysis about tourist understanding in \ref{ssec:sent_role} --- since tourist intents are highly diverse, SLU needs to extract as much information as possible from both speaker roles to help understanding.
In addition, it is reasonable that tourist understanding focuses on guide history a little bit more than tourist history under a dialogue scenario. 

For guide understanding, the tourist and the guide attention weights are 0.30 and 0.70 respectively. 
The remarkable difference of role-level attention can be explained by the data characteristics of DSTC4.
In the tourist and guide dialogues, it is obvious that the guides usually follow the same interactive patterns, such as suggesting a famous location first and then explaining it.
In other words, the tourist contexts do not matter a lot when understanding guide utterances, because the guide only needs to follow a pattern to interact with the tourist.
This also explains why guide understanding always achieves higher performance compared to the tourist part in our experiments.

To further verify whether the learned attention focuses on the right part, we conduct a similar experiment as the row (d) in Table~\ref{tab:res}.
Instead of using preceding utterances from both speakers as history information,
when understanding tourist utterances, we use only tourist history; likewise, guide understanding only takes guide history into account.
The performance of tourist understanding drops from 68.3\% to 65.9\%, while the performance of guide understanding remains almost the same (from 74.4\% to 73.8\%).
The trend proves that the content-aware attention mechanism is capable of focusing on the correct part for better understanding.
In the future, we would like to investigate whether the proposed model can generalize to the scenarios with more than two speaker roles.

\subsection{Overall Results}
From the above experimental results, the proposed time-aware attention model significantly improves the performance, and sentence-level and role-level attention has different capacities of improving understanding performance, where sentence-level attention is useful even though we do not consider content-aware information, and role-level attention is more effective when combining both content-aware and time-aware attention.

In sum, the best results are time-aware attention with sentence-level attention (row (i)) and content-aware and time-aware attention with role-level attention (row (m)), reaching higher than 70\% and 77\% on F-measure for tourist understanding and guide understanding respectively.
Therefore, the proposed attention mechanisms are demonstrated to be effective for improving SLU in such complex human-human conversations.

\section{Conclusion}
This paper proposes an end-to-end attentional role-based contextual model that automatically learns speaker-specific contextual encoding and investigates various content-aware and time-aware attention mechanisms on it.
Experiments on a benchmark multi-domain human-human dialogue dataset show that the time-aware and role-level attention mechanisms provide additional cues to guide the model to focus on the salient contexts, and achieve better performance on spoken language understanding for both speaker roles. 
Moreover, the proposed time-aware and role-level attention mechanisms are easily extendable to multi-party conversations, and we leave the extension of other variants for the future work.


\section{Acknowledgements}
We would like to thank reviewers for their insightful comments on the paper.
The authors are supported by the Ministry of Science and Technology of Taiwan and MediaTek Inc..

\bibliographystyle{IEEEbib}
\bibliography{strings,refs}

\begin{thebibliography}{10}

\bibitem{wen2017network}
Tsung-Hsien Wen, Milica Gasic, Nikola Mrksic, Lina~M Rojas-Barahona, Pei-Hao
  Su, Stefan Ultes, David Vandyke, and Steve Young,
\newblock ``A network-based end-to-end trainable task-oriented dialogue
  system,''
\newblock in {\em Proceedings of EACL}, 2017, pp. 438--449.

\bibitem{bordes2017learning}
Antoine Bordes, Y-Lan Boureau, and Jason Weston,
\newblock ``Learning end-to-end goal-oriented dialog,''
\newblock in {\em Proceedings of ICLR}, 2017.

\bibitem{dhingra2017towards}
Bhuwan Dhingra, Lihong Li, Xiujun Li, Jianfeng Gao, Yun-Nung Chen, Faisal
  Ahmed, and Li~Deng,
\newblock ``Towards end-to-end reinforcement learning of dialogue agents for
  information access,''
\newblock in {\em Proceedings of ACL}, 2017, pp. 484--495.

\bibitem{li2017end}
Xiujun Li, Yun-Nung Chen, Lihong Li, Jianfeng Gao, and Asli Celikyilmaz,
\newblock ``End-to-end task-completion neural dialogue systems,''
\newblock in {\em Proceedings of The 8th International Joint Conference on
  Natural Language Processing}, 2017.

\bibitem{tur2011spoken}
Gokhan Tur and Renato De~Mori,
\newblock {\em Spoken language understanding: Systems for extracting semantic
  information from speech},
\newblock John Wiley \& Sons, 2011.

\bibitem{hakkani2016multi}
Dilek Hakkani-T{\"u}r, G{\"o}khan T{\"u}r, Asli Celikyilmaz, Yun-Nung Chen,
  Jianfeng Gao, Li~Deng, and Ye-Yi Wang,
\newblock ``Multi-domain joint semantic frame parsing using bi-directional
  rnn-lstm.,''
\newblock in {\em Proceedings of INTERSPEECH}, 2016, pp. 715--719.

\bibitem{chen2016knowledge}
Yun-Nung Chen, Dilek Hakkani-Tur, Gokhan Tur, Asli Celikyilmaz, Jianfeng Gao,
  and Li~Deng,
\newblock ``Knowledge as a teacher: Knowledge-guided structural attention
  networks,''
\newblock {\em arXiv preprint arXiv:1609.03286}, 2016.

\bibitem{chen2016syntax}
Yun-Nung Chen, Dilek Hakanni-T{\"u}r, Gokhan Tur, Asli Celikyilmaz, Jianfeng
  Guo, and Li~Deng,
\newblock ``Syntax or semantics? knowledge-guided joint semantic frame
  parsing,''
\newblock in {\em Proceedings of 2016 IEEE Spoken Language Technology
  Workshop}, 2016, pp. 348--355.

\bibitem{wang2016learning}
Zhangyang Wang, Yingzhen Yang, Shiyu Chang, Qing Ling, and Thomas~S Huang,
\newblock ``Learning a deep l∞ encoder for hashing,''
\newblock in {\em Proceedings of IJCAI}, 2016, pp. 2174--2180.

\bibitem{bhargava2013easy}
Anshuman Bhargava, Asli Celikyilmaz, Dilek Hakkani-Tur, and Ruhi Sarikaya,
\newblock ``Easy contextual intent prediction and slot detection,''
\newblock in {\em Proceedings of ICASSP}, 2013, pp. 8337--8341.

\bibitem{xu2014contextual}
Puyang Xu and Ruhi Sarikaya,
\newblock ``Contextual domain classification in spoken language understanding
  systems using recurrent neural network,''
\newblock in {\em Proceedings of ICASSP}, 2014, pp. 136--140.

\bibitem{chen2015leveraging}
Yun-Nung Chen, Ming Sun, Alexander~I. Rudnicky, and Anatole Gershman,
\newblock ``Leveraging behavioral patterns of mobile applications for
  personalized spoken language understanding,''
\newblock in {\em Proceedings of ICMI}, 2015, pp. 83--86.

\bibitem{sun2016an}
Ming Sun, Yun-Nung Chen, and Alexander~I. Rudnicky,
\newblock ``An intelligent assistant for high-level task understanding,''
\newblock in {\em Proceedings of IUI}, 2016, pp. 169--174.

\bibitem{shi2015contextual}
Yangyang Shi, Kaisheng Yao, Hu~Chen, Yi-Cheng Pan, Mei-Yuh Hwang, and Baolin
  Peng,
\newblock ``Contextual spoken language understanding using recurrent neural
  networks,''
\newblock in {\em Proceedings of ICASSP}, 2015, pp. 5271--5275.

\bibitem{weston2015memory}
Jason Weston, Sumit Chopra, and Antoine Bordesa,
\newblock ``Memory networks,''
\newblock in {\em Proceedings of ICLR}, 2015.

\bibitem{chen2016end}
Yun-Nung Chen, Dilek Hakkani-T{\"u}r, G{\"o}khan T{\"u}r, Jianfeng Gao, and
  Li~Deng,
\newblock ``End-to-end memory networks with knowledge carryover for multi-turn
  spoken language understanding.,''
\newblock in {\em Proceedings of INTERSPEECH}, 2016, pp. 3245--3249.

\bibitem{kim2016fourth}
Seokhwan Kim, Luis~Fernando D’Haro, Rafael~E Banchs, Jason~D Williams, and
  Matthew Henderson,
\newblock ``The fourth dialog state tracking challenge,''
\newblock in {\em Proceedings of IWSDS}, 2016.

\bibitem{chi2017speaker}
Ta-Chung Chi, Po-Chun Chen, Shang-Yu Su, and Yun-Nung Chen,
\newblock ``Speaker role contextual modeling for language understanding and
  dialogue policy learning,''
\newblock in {\em Proceedings of IJCNLP}, 2017.

\bibitem{bahdanau2014neural}
Dzmitry Bahdanau, Kyunghyun Cho, and Yoshua Bengio,
\newblock ``Neural machine translation by jointly learning to align and
  translate,''
\newblock {\em arXiv preprint arXiv:1409.0473}, 2014.

\bibitem{xu2015show}
Kelvin Xu, Jimmy Ba, Ryan Kiros, Kyunghyun Cho, Aaron Courville, Ruslan
  Salakhudinov, Rich Zemel, and Yoshua Bengio,
\newblock ``Show, attend and tell: Neural image caption generation with visual
  attention,''
\newblock in {\em Proceedings of ICML}, 2015, pp. 2048--2057.

\bibitem{schuster1997bidirectional}
Mike Schuster and Kuldip~K Paliwal,
\newblock ``Bidirectional recurrent neural networks,''
\newblock {\em IEEE Transactions on Signal Processing}, vol. 45, no. 11, pp.
  2673--2681, 1997.

\bibitem{yang2017end}
Xuesong Yang, Yun-Nung Chen, Dilek Hakkani-T{\"u}r, Paul Crook, Xiujun Li,
  Jianfeng Gao, and Li~Deng,
\newblock ``End-to-end joint learning of natural language understanding and
  dialogue manager,''
\newblock in {\em Proceedings of ICASSP}, 2017, pp. 5690--5694.

\bibitem{sukhbaatar2015end}
Sainbayar Sukhbaatar, Jason Weston, Rob Fergus, et~al.,
\newblock ``End-to-end memory networks,''
\newblock in {\em Proceedings of NIPS}, 2015, pp. 2431--2439.

\bibitem{xiong2016dynamic}
Caiming Xiong, Stephen Merity, and Richard Socher,
\newblock ``Dynamic memory networks for visual and textual question
  answering,''
\newblock {\em arXiv preprint arXiv:1603.01417}, 2016.

\bibitem{pennington2014glove}
Jeffrey Pennington, Richard Socher, and Christopher~D Manning,
\newblock ``Glove: Global vectors for word representation.,''
\newblock in {\em Proceedings of EMNLP}, 2014, vol.~14, pp. 1532--1543.

\end{thebibliography}

\end{document}